\documentclass[pmlr,twocolumn,10pt]{jmlr} % W&CP article

\usepackage{booktabs}
\usepackage{siunitx}

\usepackage[switch]{lineno}

\theorembodyfont{\upshape}
\theoremheaderfont{\scshape}
\theorempostheader{:}
\theoremsep{\newline}

\usepackage{bm}
\usepackage{nccmath}
\usepackage{multirow}
\usepackage{caption}
\usepackage{subcaption}
\usepackage{lipsum}  
\usepackage{graphicx}
\usepackage[normalem]{ulem}
\useunder{\uline}{\ul}{}

\title{Temporally Multi-Scale Sparse Self-Attention for Physical Activity Data Imputation}

\author{
\Name{Hui Wei} \Email{huiwei@cs.umass.edu} \\
\addr University of Massachusetts Amherst, United States
\\
\Name{Maxwell A. Xu} \Email{maxxu@gatech.edu} \\
\addr Georgia Institute of Technology, United States
\\
\Name{Colin Samplawski} \Email{csamplawski@cs.umass.edu} \\
\addr University of Massachusetts Amherst, United States
\\
\Name{James M. Rehg} \Email{jrehg@illinois.edu} \\
\addr University of Illinois Urbana-Champaign, United States
\\
\Name{Santosh Kumar} \Email{santosh.kumar@memphis.edu} \\
\addr University of Memphis, United States
\\
\Name{Benjamin M. Marlin} \Email{marlin@cs.umass.edu} \\
\addr University of Massachusetts Amherst, United States
}

\begin{document}

\maketitle

%%%% ABSTRACT %%%%
\begin{abstract}
  Wearable sensors enable health researchers to continuously collect data pertaining to the physiological state of individuals in real-world settings. However, such data can be subject to extensive missingness due to a complex combination of factors. In this work, we study the problem of imputation of missing step count data, one of the most ubiquitous forms of wearable sensor data. We construct a novel and large scale data set consisting of a training set with over 3 million hourly step count observations and a test set with over 2.5 million hourly step count observations. We propose a domain knowledge-informed sparse self-attention model for this task that captures the temporal multi-scale nature of step-count data. We assess the performance of the model relative to baselines and conduct ablation studies to verify our specific model designs.
\end{abstract}

%%%% DATA CODE AVAILABILITY %%%%
\paragraph*{Data and Code Availability}
This paper uses the \textit{All of Us} dataset\footnote{\url{https://www.researchallofus.org/}}, which is publicly available upon registration. Data processing and modeling code is available at \url{https://github.com/reml-lab/allofus-imputation}.

\paragraph*{Institutional Review Board (IRB)}
This research does not require IRB approval.

%%%% INTRODUCTION %%%%
\section{Introduction} \label{sec:intro}

Step count data collected by smart watches and activity trackers is one of the most ubiquitous forms of wearable sensor data. These data have the potential to provide valuable and detailed information about physical activity patterns and their relationship to other facets of health over long time spans. These data also have the potential to provide valuable contextual information for just-in-time adaptive interventions that target improving levels of physical activity or deceasing sedentary behavior \citep{Rehg2017mhealth,spruijt2022advancing}. However, wearable sensor data are subject to complex missingness patterns that arise from a variety of causes including device non-wear, insecure device attachment and devices running out of battery \citep{tackney2022missing,lin2020filling, Rahman2017mdebugger}. 

Importantly, these missingness issues can hinder the utility of wearable sensor data to support both improved understanding of health behaviors and to provide actionable contexts in the case of adaptive interventions. Indeed, the presence of missing step count data is a problem for traditional statistical analyses that aim to relate physical activity levels to other health events and to the effect of interventions \citep{master2022association, hall2020systematic}.  Missing step count data is also a problem when practitioners seek to use these data as inputs to common supervised and unsupervised models that require complete data as input \citep{papathomas2021machine}, as well as when step count data is used in the reward function for reinforcement learning-based adaptive interventions \citep{liao2020personalized, zhou2018personalizing}. 

In this paper, we consider the problem of imputing missing step count data at the hourly level. This problem has a number of significant challenges due to the presence of high variability in patterns of physical activity both through time for a single person and between different people. This variability can be attributed to a collection of factors that are exogenous to step count data itself including an individual's levels of restedness and business, environmental factors such as weather and temperature, changes in daily routine, seasonal effects, onset and recovery from illness and other major life events. To make progress on these challenges necessitate both carefully designed, domain-informed models and the availability of large-scale step count datasets. 

To address the need for a large-scale data set, we curate a training set consisting of hourly step count data from 100 individuals. The average step count time series length is over $50,000$ hourly observations per person in the training set yielding a total of over 3 million hourly step count observations. We curate a test set consisting of data from 500 individuals including over 2.5 million observed hourly step count instances. This data set is based on minute-level Fitbit step count data collected as part of the \textit{All of Us} research project \citep{mapes2020diversity, mayo2023all}. The \textit{All of Us} data set is freely available to registered researchers\footnote{\url{https://www.researchallofus.org}}. 

To address the modeling challenges, we introduce a novel sparse self-attention model inspired by the transformer architecture \citep{vaswani2017attention}. The proposed model uses sparse attention to handle the quadratic complexity of the standard dense self-attention mechanism, which is not practical given long time series as input.  Importantly, the sparse self-attention mechanism is designed to be temporally multi-scale in order to capture diurnal, weekly, and longer time-scale correlations. The specific design used is informed by an analysis of hourly step count autocorrelations. Finally, we design an input feature representation that combines a  time encoding (hour of day, day of week) with a temporally local activity pattern representation.

We compare our proposed model to a broad set of prior models and approaches including a convolutional denoising autoencoder that achieved state-of-the-art performance on missing data imputation in actigraphy data \citep{jang2020deep}. The results show that our model achieves statistically significant improvements in average predictive performance relative to the prior approaches considered at the $p<0.05$ level. We further break down performance by missing data rate and ground truth step count ranges. Finally, we visualize attention weights and relative time encodings to investigate what the proposed model learns and conduct an ablation study of the key components of the proposed model.

We begin by discussing related work in Section \ref{sec:relatedworks}, and then describe our dataset in Section \ref{sec:data}. We describe our proposed self-attention imputation model in Section \ref{sec:method}. In Section \ref{sec:exp}, we describe our experimental methods and in Section \ref{sec:results}, we report our experimental results.

%%%% RELATED WORK %%%%
\section{Related Work} \label{sec:relatedworks}

In this section, we briefly review general missing data imputation methods for time series, prior work on sparse self-attention, and prior work specifically on step count imputation models.

\noindent\textbf{Imputation Methods for Time Series}  
The missing data imputation problem has been intensively studied in both statistics \citep{little2019statistical} and machine learning \citep{emmanuel2021survey, gond2021survey}. Commonly used baseline  methods include mean imputation \citep{emmanuel2021survey}, regression imputation \citep{little1992regression}, $k$-nearest neighbors ($k$NN) imputation, and multiple imputation by chained Equations (MICE) \citep{little2019statistical, azur2011multiple}. Both regression imputation and MICE are model-based approaches that aim to impute missing values as functions of observed variables while ($k$NN) is a non-parametric approach. 

More recently, the machine learning community has focused on neural network-based imputation methods for time series including the use of recurrent neural networks (RNNs) \citep{hochreiter1997long, cho2014learning} and generative adversarial networks (GAN) \citep{goodfellow2014generative}. \cite{che2018recurrent} introduced the gated recurrent unit with decay (GRU-D) model for irregularly sampled and incomplete time series data, which takes into account missingness patterns and time lags between consecutive observations \citep{cho2014learning}. In the imputation setting, uni-directional RNN models like GRU-D are typically outperformed by bi-directional RNN models such as the M-RNN \citep{yoon2018estimating} and BRITS \citep{cao2018brits}.

While basic GAN models for fully observed data require only a generator and discriminator, training these models using partially observed data can require architectural or training modifications. \cite{luo2018multivariate} trained a GAN model in two stages to select noise capable of generating samples most similar to the original values. \cite{luo2019e2gan} proposed $E^{2}\text{GAN}$, which uses an autoencoder architecture as the generator, enabling end-to-end training and eliminating the need for two-stage training. Additionally, \cite{miao2021generative} (SSGAN) introduced a temporal remainder matrix as a hint to the discriminator to facilitate training. SSGAN also used time series class labels to guide the generation procedure with USGAN provising an non-class supervised alternative. 

In this work, we focus on self-attention-based imputation models trained using empirical risk minimization (ERM). Self-attention based models are well-known to have improved parallelization compared to RNN-based models \citep{martin2018parallelizing}. The use of ERM-based training (e.g., prediction loss minimization) avoids  stability issues inherent to current GAN-based model training algorithms \citep{sinha2020top, arjovsky2017towards}. Our primary modeling contribution focuses on making self-attention models computationally efficient for long time series of step counts using sparsity. We discuss prior work on sparse self-attention in the next section. 

\noindent\textbf{Sparse Self-Attention} Many methods have attempted to address the quadratic complexity of self-attention computations using sparsity \citep{tai2023effecienttrans}. For instance, the vision transformer \citep{dosovitskiy2021image} and Swin transformer \citep{liu2021swin} apply self-attention on non-overlapping patches in an image. The sparse transformer \citep{child2019generating} and axial transformer \citep{ho2019axial} separate the full attention map into several attention steps using multiple attention heads. Several authors have also investigated learnable sparsity mechanisms. Deformable DETR \citep{zhu2021deformable}, Reformer \citep{kitaev2020reformer} and Routing Transformer \citep{roy2021efficient} retrieve the most relevant keys for each query using learnable sampling functions, locality sensitivity hashing, and $k$-means, respectively. The drawback of these approaches is that they typically require higher training times. Our proposed model uses a fixed, multi-timescale sparsity pattern that is designed specifically for step count data. 

\noindent\textbf{Step Count Imputation} \cite{pires2020improving} used $k$NN imputation for step count data collected from accelerometers and magnetometers. \cite{tackney2023multiple} employed multiple imputation methods combined with both parametric (e.g., regression imputation) and non-parametric approaches (e.g., hot deck imputation) to impute missing daily and hourly step count data. \cite{ae2018missing} proposed a zero-inflated Poisson regression model to handle zero step count intervals more effectively. \cite{jang2020deep} used a convolutional denoising autoencoder architecture that exhibited superior performance compared to multiple other approaches including mean imputation, Bayesian regression and the zero-inflated model by \cite{ae2018missing}. In this work, we focus on model-based single imputation and compare to a wide range of baseline and current stat-of-the art approaches on large-scale data.

%%%% DATA DESCRIPTION %%%%
\section{Data Set Development} \label{sec:data}

In this section, we describe the curation and prepossessing methods we apply to develop the data set used in our experiments. Flowcharts summarizing our methods are provided in Appendix \ref{sec:curation_preprocess}. 

\noindent\textbf{Data Set Extraction} Our data set is derived from the \textit{All of Us} research program Registered Tier v6 data set  \citep{mayo2023all}. \textit{All of Us} is an NIH-funded research cohort with an enrollment target of one million people from across the U.S. The v6 data set includes minute-level step count and heart rate data collected using Fitibt devices from 11,520 adult participants. While the \textit{All of Us} research program directly provides daily step count summaries derived from these data, we focus on the finer-grained problem of imputing missing step count data at the hourly level. This timescale is highly relevant for applications like the analysis of adaptive interventions that need access to finer-grained step count data to assess the proximal effects of actions. Further, due to devices running out of battery during the day and temporary device non-wear, the base data set contains substantial partial within-day missingness that can be usefully imputed to support a variety of downstream analyses.

We begin by rolling-up the minute-level Fitibit time series for each participant  into an hourly time series. We use one-hour long blocks aligned with the hours of the day. Each block is represented by the total observed steps within that hour, the average heart rate within that hour, and the number of minutes of observed data (the wear time) within that hour. The range of minutes of wear time for each hourly block is 0-60. We define hourly blocks with zero minutes of wear time as \emph{missing}, and hourly blocks with at least one minute of wear time as \emph{observed} (our modeling approach will specifically account for observed hourly blocks with different wear time).    

Imputation model training requires holding out \emph{observed} data to use as prediction targets thus increasing the amount of missing data seen by models during training. Also, learning on more complete data makes it easier for models to identify appropriate physical activity structure in the data. Therefore, we form a training set of individuals with low to moderate levels of natural missing data. Specifically, we select for the training set the $100$ participants with the most observed hourly blocks among those with at least one $180$ day long segment of step count data containing no run of missing hourly data longer than three days. The resulting training data set consists of over 3 million observed hourly blocks with an average time series length of over $50,000$ hours per training set participant. 

Since many participants do not wear their devices between 11:00pm and 5:00am and the observed step count data for those who do is almost always 0 (presumably due to sleep), we focus on predicting step counts in the interval of 6:00am to 10:00pm (we use data outside of this range as part of the feature representation for some models). The maximum missing data rate among the training participants is 20\% within the 6:00am to 10:00pm time frame. Appendix \ref{sec:comp_cohort_allofus} provides comparisons between the 100 participants in our training cohort and all 11,520 participants in the \textit{All of Us} Fitbit dataset.

To form a test set, we first exclude the training participants. Next, we select a total of 100 participants for each of five missing data level bins [0\%, 20\%), [20\%, 40\%), [40\%, 60\%), [60\%, 80\%), and [80\%, 100\%). We again assess missing data within the 6:00am to 10:00pm time frame. For the [0\%, 20\%) bin, we apply the same filtering criteria as for the training set and select 100 participants at random from those meeting the criteria. For the remaining bins, we select participants at random with no additional criteria. This yields a total of 500 test participants with a total of approximately 2.5 million observed hourly blocks.

\noindent\textbf{Data Set Pre-Processing} Once the data set is extracted, we apply several pre-processing steps. First, to deal with partially observed hourly blocks, the model that we construct uses step rates as features instead of step counts. The step rate associated with an hourly block is defined as the observed step count divided by the observed wear time. When making predictions for observed hourly blocks, the model predicts a step rate, but the loss is computed between the observed step count and a predicted step count formed by combining the predicted step rate with the observed wear time. 

Further, we use the mean and standard deviation of each participant's step rate and heart rate data 
(ignoring outliers beyond the 99.9\% percentile) to compute statistics for z-normalization \citep{ulyanov2016instance} of step rates and heart rates. This z-normalization step is applied separately to each participant's data to provide an initial layer of robustness to between-person variability. In order to enable vectorized computations over time series with missing data, we use zero as a placeholder for missing data values and use an auxiliary response indicator time series to maintain information about which blocks are missing and which are observed.

Finally, the raw Fitbit time series provided by the \textit{All of Us} research program were shifted by a randomly selected number of days for each participant as part of a set of privacy preserving transformations. In order to enable models to learn common behavior patterns with respect to day of the week, we select a reference participant and align all other participants to that participant by considering all shifts of between 0 and 6 days. We use similarity in average daily step counts as the alignment criteria. While we can not be certain that this process recovers the correct shift, it will decrease variability relative to the baseline of not applying this correction. 

%%%% MODEL %%%%
\section{Proposed Model}\label{sec:method}
In this section, we formally define the step count imputation problem within the multivariate context, and introduce our temporally multi-scale sparse self-attention model architecture.

%\subsection{Problem Definition}
\noindent\textbf{Problem Definition} We denote by $\mathcal{D}=\{\mathbf{C}^{(n)}_{l,t}|n=1,\ldots,N, l=1,\ldots,L, t=1,\dots,T_{n} \}$ a dataset of $N$ participants, where each participant is represented by a multivariate time series, $\mathbf{C}^{(n)} \in \mathbb{R}^{L \times T_{n}}$ with $L$ features and $T_{n}$ hourly blocks. $T_{n}$ varies across participants, while the number of features $L$ is constant. In our case, the base features associated with each hourly block include step count, step rate, heart rate, day of the week, hour of the day and minutes of wear time. 
%We index the step count as the first feature, $\mathbf{C}_{1,:}^{(n)}$. 
When considering data from a single participant, we drop the $(n)$ superscript for brevity. 

For each hourly block $t$, we define the response indicator $r_t$ as shown in \equationref{equ:res_ind} to indicate if the participant's Fitbit data at a given hourly block is observed (i.e. with at least one minute of wear time). We let $\mathbf{C}_{w,t}$ be the wear time. While heart rates may contain missing values, our focus in this study is not on imputing them. We also note that the hour of the day, day of the week and wear time itself are always completely observed.

\begin{equation}
\small
\label{equ:res_ind}
r_{t} = \begin{cases}
            1 &  \mathbf{C}_{w,t}>0 \\
            0 & \text{otherwise}
          \end{cases}
\end{equation}

We let $\mathbf{C}_{s,t}$ be the step count at time at time $t$. The problem is thus to impute $\mathbf{C}_{s,t}$ when $r_{t} = 0$ from the observed data. This includes observed Fitbit data from other time steps as well as other observed data at time step $t$. Crucially, we can only train and assess imputation models on \emph{originally observed} hourly blocks in the dataset since they have ground-truth Fitbit data values. Thus, instead of imputing \emph{originally missing} hourly blocks that do not have ground-truth values, we hold out hourly blocks with \emph{observed} values, consider them as ``artificially missing", then use models to predict their original observed values.

\noindent\textbf{Model Overview} We propose a model architecture based on dot-product self-attention \citep{vaswani2017attention}. As noted previously, the standard transformer architecture uses dense self-attention, which has quadratic cost in the length of an input time series. This is highly prohibitive for long time series. Indeed, our training data set has an average time series length of $50,000$ hours per participant. This is longer than the context window used in some versions of GPT-4 \citep{achiam2023gpt}. Thus, the first key component of our proposed architecture is the design of a sparse self-attention structure for step count imputation. Based on domain knowledge combined with data analysis, we propose a self attention mechanism based on a multi-timescale context window. The second key component of the architecture is the feature representation. While transformer models applied to text data typically use a base token embedding computed from fully observed data, we require an input representation that is specific to this task. We propose a local activity profile representation (LAPR) that represents hourly blocks with a temporally local window of activity data. 

\noindent\textbf{Sparse Self-Attention}  In order to construct a  self-attention-based model for long time series, we need to drastically reduce the number of hourly blocks attended to by each query for each missing hourly block. To begin, let $\mathcal{T}=\{1,\ldots,T\}$ be the set of all the hourly blocks from a given participant and $|\mathcal{T}|$ be the size of this set. We define the set $\mathcal{A}^{(t)}\subseteq \mathcal{T}$ to be a sub-set of hourly blocks that a query at time $t$ is allowed to attend to. For improved computational efficiency, we require $|\mathcal{A}^{(t)}| \ll |\mathcal{T}|$ for all $t$. However, in the missing data context, even if a time point $t$ is allowed to attend to a time point $t'$, time point $t'$ may not have observed data. We define a mask function $m(t, t')$ in \equationref{equ:mask_theta} that indicates both whether time point $t$ can attend to time point $t'$ and whether time point $t'$ is observed.

\begin{align}
\label{equ:mask_theta}
m(t, t') &= \begin{cases}
                  1 & t' \in \mathcal{A}^{(t)} \text{ and } r_{t'}=1  \\
                  0 & \text{otherwise}
            \end{cases}
\end{align}

The key question is then how to define the self-attention sets $\mathcal{A}^{(t)}$. Based on domain knowledge, we expect that hourly blocks $t'$ that are close in time to a given target hourly block $t$ will carry information useful to make predictions at time $t$. However, we also expect that hourly blocks $t'$ corresponding to the same hour of the day as a target block $t$ on nearby days may carry information useful to make predictions at time $t$. Similarly, we expect that hourly blocks $t'$ corresponding to the same hour of the day and the same day of the week for nearby weeks may also carry information useful to make predictions at time $t$. 

\begin{figure}[t]
\includegraphics[width=\linewidth]{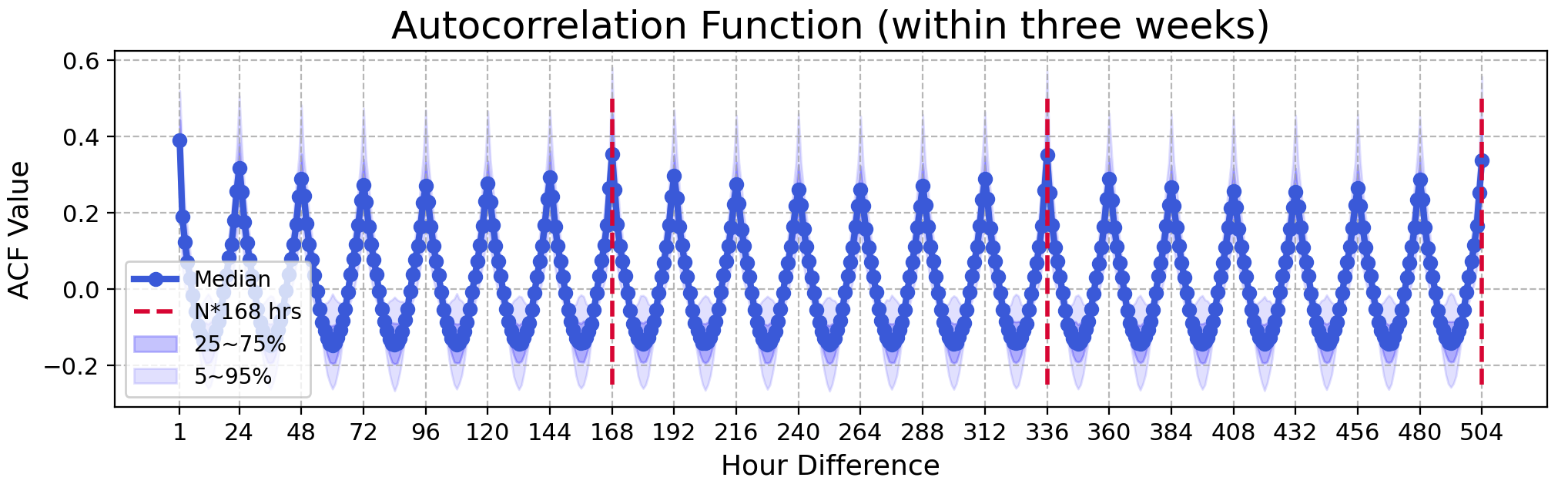}
\caption{Autocorrelation function (ACF) over all the participants of $\Delta t = 1, \ldots, 504$ hrs (within three weeks). \textbf{Blue line}: median ACF, \textbf{Red line}: $\Delta t=168 \times N$ hrs (i.e. $N$ weeks).}
\label{fig:acf}
\end{figure}

\begin{figure}[t]
\centering
\includegraphics[height=3cm, width=\columnwidth, keepaspectratio] {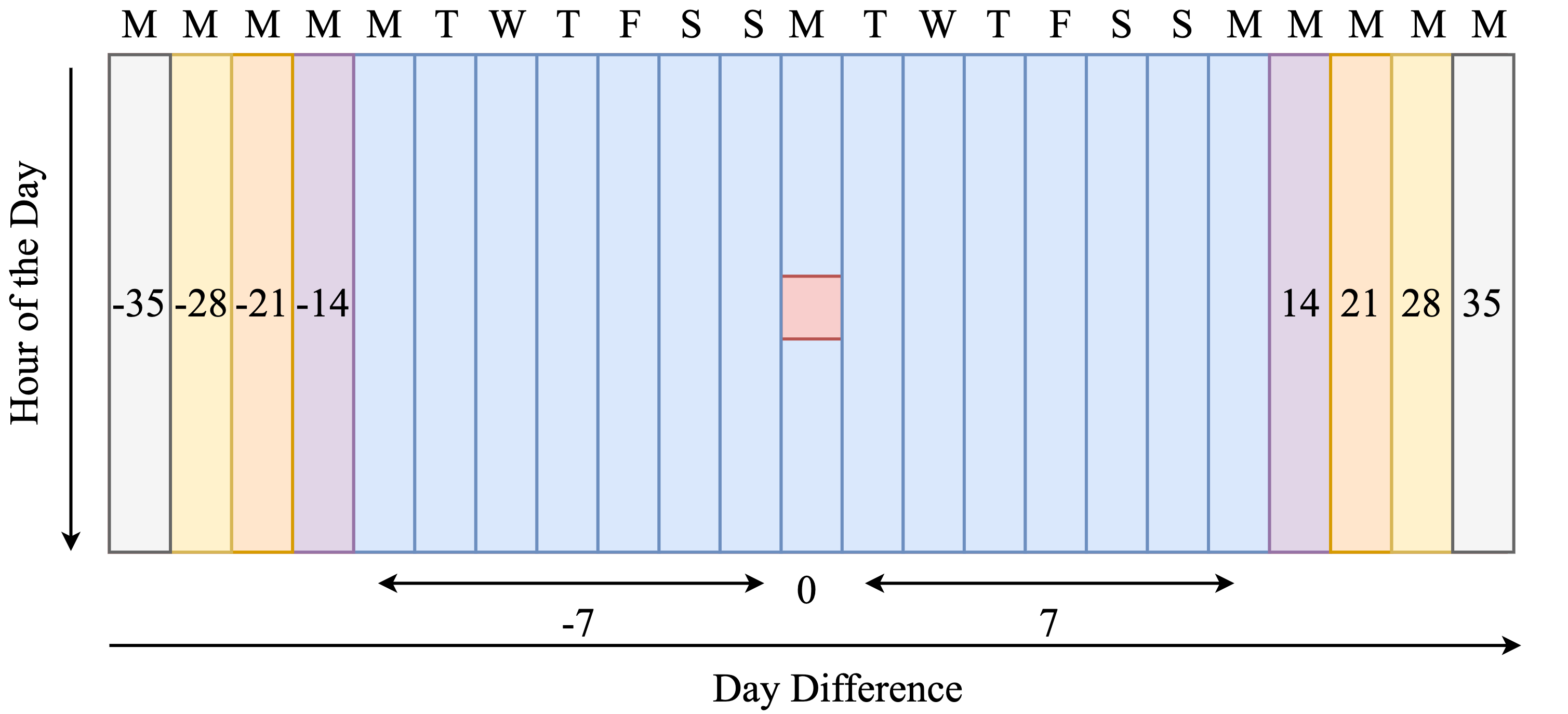}
\caption{Multi-timescale context window. The missing hourly block is at the center and indicated as red. Numbers are day differences between each day and the center day (i.e. difference is 0) which contains the missing hourly block. Letters indicate the day of the week for each day. The center day is Monday in this example.} 
\label{fig:context_window_discontinuous}
\end{figure}

In Figure \ref{fig:acf}, we present the hourly step count autocorrelation function for our data set to confirm these expectations. First, we can see that the autocorrelation is highest for the smallest time lags indicating high correlation between nearby hourly blocks. However, we can also see strong correlations at time lags of 24 hours (1 day) and 168 hours (1 week). This confirms our expectations regarding the general correlation structure of the data. 

Based on these observations, we propose the multi-time scale context window shown in Figure \ref{fig:context_window_discontinuous} as our sparse self-attention set $\mathcal{A}(t)$. Letting $d$ be the day number of the target hourly block $t$, the context window includes data from days $d$ to day $d\pm 7$ as well as $d\pm 7k$ for $k\in\{2,3,4,5\}$. Given that time points $t'$ with hour of the day closer to the target hour $t$ have higher correlations, we limit the context window to include time points $t'$ with hours of the day close to that of $t$. Letting $h$ be the hour of the day for the target hourly block $t$, the context window includes hours $h$ to $h\pm 4$. Of course, the center of the context window, which corresponds to the target hourly block $t$, is not included in the sparse self-attention set. This yields a total self-attention set size of $(2\times4+1)(2\times(7+4)+1)-1 = 206$.

\noindent\textbf{Feature Representation} Individual hourly blocks are featurized in terms of step count, step rate, average heart rate, wear time minutes, hour of the day, and day of the week. However, the target hourly block has its Fitbit features (i.e. step count, step rate and heart rate) unobserved. A self-attention computation based on comparing the observed features of the target hourly block to the corresponding features in blocks in the self-attention set would thus be limited to expressing similarity based on hour of the day and day of the week. 

To overcome this problem, we augment the representation of an hourly block's step rate data using a window of activity data from $t-W$ to $t+W$. We refer to this as the ``local activity profile" representation (LAPR) of an hourly block. It allows for learning much richer notions of similarity between hourly blocks within the multi-scale context windows based on comparing their local activity profiles. As described in Section \ref{sec:exp}, missing values in the LAPR feature representation are themselves imputed using a baseline approach.

\noindent\textbf{Proposed Model} The proposed model is summarized in the equations below. $s_{t}$ is the predicted step rate at time $t$. $a_{tt'}$ is the attention weight from hourly block $t$ to hourly block $t'$. $m(t,t')$ is the sparse attention mask function defined in \equationref{equ:mask_theta}. The sparse attention mask ensures that the attention weight is $0$ for time points $t'$ that are not included in the sparse self attention context window as well as points $t'$ with missing Fitbit data.

%\useshortskip
\begin{align}
\label{equ:attn_avg}
\small
s_{t} &= \sum_{t' \neq t} a_{tt'} v_{t'}\\
%
% \useshortskip
\label{equ:attn_weight}
\small
a_{tt'} &= \frac{m(t,t')\exp(\mathbf{q}_{t}^\top \mathbf{k}_{t'} + \bm{\theta}_{\mathtt{I}(t,t')})}
                {\sum_{u \neq t} m(t,u)\exp(\mathbf{q}_{t}^\top \mathbf{k}_{u} + \bm{\theta}_{\mathtt{I}(t,u)})}
\end{align}

The primary components of the self-attention computation are the value $v_{t'}$, the query vector $\mathbf{q}_{t}$, the key vector $\mathbf{k}_{t'}$ and the relative time embedding $\bm{\theta}_{\mathtt{I}(t,t')}$. The value $v_{t'}$, the query vector $\mathbf{q}_{t}$ and the key vector $\mathbf{k}_{t'}$ are produced using distinct neural network-based transformations of the input features for their respective time points. To begin, the local activity profile representation (LAPR) is processed through an encoder network $Conv \to LayerNorm \to ReLU \to Average Pool$. This encoder extracts more abstract features and also prevents the overfitting problem by lowering the input dimension. The output of the encoder is then concatenated with the other available features. For the key and the value, this includes the hour of the day and day of week features as well as the Fitbit features of that specific time point. For the query, the Fitbit features for the target time point $t$ are not observed, so the LAPR is concatenated with the hour and day features only. We use a one-hot encoding representation for the hour and day features. The resulting representation is projected through linear layers to produce the final query, key and value representations. 

To encode information based on the time difference between the target hourly block $t$ and another block $t'$, a relative time encoding $\bm{\theta}_{\mathtt{I}(t,t')}$ is employed. Essentially, the model provides an attention bias parameter for each position in the context window. This allows the model to learn that some relative positions in the context window are valuable to attend to regardless of the similarity in feature values at those relative locations for a particular instance. The function $\bm{\theta}_{\mathtt{I}(t,t')}$ returns the value of the relative time encoding bias parameter for time point $t'$ in the context window centered at time $t$. If $t'$ falls outside of the context window, this function returns $0$. 

\noindent\textbf{Loss Function and Training} The output of the model is an unconstrained hourly step rate. We convert the hourly step rate to a step count using the transformation $\mathbf{C}_{w,t}\cdot \min(1.5\cdot s_{max}, \max(0, s_{t}))$ where $\mathbf{C}_{w,t}$ is the observed wear time for time $t$, and $s_{max}$ is the maximum training set step rate observed for the participant. This ensures that the step count is always non-negative and clips the maximum predicted step rate to avoid predicting outlying values. 

We use mean absolute error (MAE) between true and predicted step counts as the loss function during model training. We use a stochastic gradient descent-based training approach where each batch contains instances sampled from different participants. We compute the MAE with equal weight on all samples in the batch. Additional hyper-parameter optimization and training details can be found in Appendix \ref{sec:model_setups}.

%%%% EXPERIMENTS %%%%

\begin{table*}[t]
\centering
\resizebox{\textwidth}{!}{%
\begin{tabular}{@{}llllllll@{}}
\toprule
            &                       & \multicolumn{5}{c}{\textbf{Missing Rate}}                                                   & \multicolumn{1}{c}{} \\ \cmidrule(lr){3-7}
\textbf{Method Category} &
  \textbf{Method} &
  \multicolumn{1}{c}{\textbf{{[}0\%, 20\%)}} &
  \multicolumn{1}{c}{\textbf{{[}20\%, 40\%)}} &
  \multicolumn{1}{c}{\textbf{{[}40\%, 60\%)}} &
  \multicolumn{1}{c}{\textbf{{[}60\%, 80\%)}} &
  \multicolumn{1}{c}{\textbf{{[}80\%, 100\%)}} &
  \multicolumn{1}{c}{\textbf{Overall}} \\ \midrule
            & Zero Fill & 474.03 ± 34.43 & 408.37 ± 28.70 & 384.86 ± 33.68 & 440.94 ± 43.45 & 422.73 ± 58.58 & 395.79 ± 18.08 \\
            & Forward Fill & 416.22 ± 32.09 & 373.31 ± 29.22 & 329.69 ± 30.76 & 381.71 ± 40.83 & 352.61 ± 47.99 & 351.87 ± 16.05 \\
            & Backward Fill & 411.61 ± 30.16 & 368.52 ± 28.74 & 324.53 ± 28.92 & 373.68 ± 39.72 & 343.54 ± 49.93 & 346.46 ± 15.98  \\
\multirow{-4}{*}{Basic Fill} &
  Avg. F+B Fill &
  {\color[HTML]{3531FF} {\ul 350.79 ± 26.43}} &
  {\color[HTML]{3531FF} {\ul 316.96 ± 25.38}} &
  {\color[HTML]{3531FF} {\ul 278.37 ± 25.75}} &
  {\color[HTML]{3531FF} {\ul 321.10 ± 34.29}} &
  {\color[HTML]{3531FF} {\ul 295.56 ± 41.23}} &
  {\color[HTML]{3531FF} {\ul 306.80 ± 13.86}} \\ \midrule
            & Participant & 414.57 ± 32.10 & 375.79 ± 27.87 & 329.57 ± 27.44 & 385.16 ± 37.74 & 367.59 ± 54.93 & 356.03 ± 16.32 \\
            & Day of Week & 411.82 ± 31.97 & 373.30 ± 27.70 & 327.20 ± 27.36 & 380.14 ± 37.34 & 355.83 ± 51.14 & 351.70 ± 15.74 \\
            & Hour of Day  & 373.93 ± 29.07 & 343.80 ± 25.07 & 303.03 ± 24.84 & 351.89 ± 32.28 & 311.45 ± 35.07 & 326.44 ± 12.88 \\
\multirow{-4}{*}{Micro Mean Fill} &
  DW+HD &
  {\color[HTML]{3531FF} {\ul 357.56 ± 27.79}} &
  {\color[HTML]{3531FF} {\ul 330.69 ± 24.15}} &
  {\color[HTML]{3531FF} {\ul 288.72 ± 23.93}} &
  {\color[HTML]{3531FF} {\ul 325.40 ± 29.62}} &
  {\color[HTML]{3531FF} {\ul 253.56 ± 30.15}} &
  {\color[HTML]{3531FF} {\ul 304.08 ± 12.12}} \\ \midrule
            & Participant & 416.01 ± 32.22 & 378.09 ± 27.99 & 332.59 ± 27.72 & 389.68 ± 38.32 & 370.80 ± 53.86 & 358.64 ± 16.27 \\
            & Day of Week  & 413.29 ± 32.10 & 375.67 ± 27.83 & 330.21 ± 27.64 & 384.73 ± 37.88 & 359.88 ± 50.41 & 354.50 ± 15.74  \\
            & Hour of Day   & 375.40 ± 29.19 & 346.02 ± 25.17 & 305.75 ± 25.09 & 356.46 ± 32.80 & 317.22 ± 35.80 & 329.50 ± 13.04  \\
\multirow{-4}{*}{Mean Fill} &
  DW+HD &
  {\color[HTML]{3531FF} {\ul 359.07 ± 27.90}} &
  {\color[HTML]{3531FF} {\ul 332.97 ± 24.26}} &
  {\color[HTML]{3531FF} {\ul 291.78 ± 24.21}} &
  {\color[HTML]{3531FF} {\ul 330.89 ± 29.97}} &
  {\color[HTML]{3531FF} {\ul 262.46 ± 30.73}} &
  {\color[HTML]{3531FF} {\ul 308.03 ± 12.21}} \\ \midrule
            & Participant & 369.17 ± 26.48 & 331.99 ± 23.03 & 299.41 ± 24.88 & 341.93 ± 32.48 & 323.09 ± 46.69 & 323.69 ± 14.12 \\
            & Day of Week & 366.66 ± 26.38 & 329.82 ± 22.84 & 297.14 ± 24.76 & 337.97 ± 32.19 & 314.87 ± 45.49 & 320.37 ± 13.92 \\
            & Hour of Day & 335.85 ± 24.41 & 307.19 ± 21.37 & 276.57 ± 22.96 & 316.67 ± 29.11 & 280.66 ± 32.40 & 300.62 ± 11.85  \\
\multirow{-4}{*}{Median Fill} &
  DW+HD &
  {\color[HTML]{3531FF} {\ul 322.03 ± 23.78}} &
  {\color[HTML]{3531FF} {\ul 295.09 ± 20.67}} &
  {\color[HTML]{3531FF} {\ul 262.29 ± 22.04}} &
  {\color[HTML]{3531FF} {\ul 292.01 ± 26.82}} &
  {\color[HTML]{FE0000} \textbf{230.54±27.15}} &
  {\color[HTML]{3531FF} {\ul 280.39 ± 11.18}} \\ \midrule
            & Uniform  & 332.70 ± 25.34 & 306.68 ± 24.28 & 270.34 ± 23.74 & 321.63 ± 36.07 & 295.93 ± 40.81 & 305.46 ± 13.93       \\
\multirow{-2}{*}{$k$NN} &
  Softmax &
  {\color[HTML]{3531FF} {\ul 331.37 ± 25.08}} &
  {\color[HTML]{3531FF} {\ul 305.90 ± 24.19}} &
  {\color[HTML]{3531FF} {\ul 269.58 ± 23.65}} &
  {315.26 ± 33.16} &
  {290.78 ± 37.53} &
  {\color[HTML]{3531FF} {\ul 302.58 ± 13.22}} \\ \midrule
\multirow{7}{*}{Model-based} & Iterative Imputation \citep{azur2011multiple}  & 313.23 ± 23.26 & 290.48 ± 21.32 & 260.61 ± 22.02 & 304.16 ± 28.86 & 289.20 ± 37.43 & 291.54 ± 12.29       \\
            & CNN-DAE \citep{jang2020deep}  & 317.26 ± 23.42 & 287.35 ± 21.27 & 256.22 ± 22.38 & 299.14 ± 30.66 & 284.27 ± 40.75 & 288.85 ± 12.93       \\
            & Regression Imputation \citep{little1992regression} & 307.96 ± 22.88 & 284.06 ± 20.69 & 254.82 ± 21.69 & 296.01 ± 27.76 & 282.21 ± 36.75 & 285.01 ± 12.01       \\
            & BRITS \citep{cao2018brits}   & 299.46 ± 21.59 & 275.58 ± 19.93 & 248.21 ± 21.40 & 289.82 ± 28.34 & 275.09 ± 36.46 & 277.63 ± 11.85       \\
            & USGAN \citep{miao2021generative} & 299.58 ± 21.58 & 275.58 ± 19.91 & 248.34 ± 21.43 & 289.68 ± 28.36 & 274.56 ± 36.36 & 277.55 ± 11.84       \\
             & SAITS  \citep{du2023saits} & 291.37 ± 20.80 & 269.50 ± 19.65 & 246.17 ± 21.06 & 282.62 ± 26.85 & 264.61 ± 31.32 & 270.85 ± 10.97 \\

            %& MRNN  \citep{yoon2018estimating}  & 295.30 ± 20.91 & 270.38 ± 19.66 & 243.96 ± 21.21 & 280.69 ± 27.25 & 263.42 ± 33.20 & 270.75 ± 11.27       \\
            &MRNN  \citep{yoon2018estimating} & 291.71 ± 20.48 & 269.07 ± 19.40 & 242.92 ± 21.03 & 281.18 ± 27.27 & 255.19 ± 30.73 & 268.02 ± 10.93 \\
 &
  \textbf{Sparse Self-Attention (ours)} &
  {\color[HTML]{FE0000} \textbf{285.76±20.41}} &
  {\color[HTML]{FE0000} \textbf{262.96±19.16}} &
  {\color[HTML]{FE0000} \textbf{239.01±20.58}} &
  {\color[HTML]{FE0000} \textbf{270.32±25.62}} &
  {\color[HTML]{3531FF} {\ul 250.36 ± 30.12}} &
  {\color[HTML]{FE0000} \textbf{261.68±10.62}} \\ \bottomrule
\end{tabular}%
}
\caption{Macro MAE ± 95\% CI on completely held-out test participants. There are 100 randomly sampled participants in each missing interval. The entire held-out test cohort contains 500 participants in total. {\color{red} Red}: overall best performance with statistical significance. {\color{blue} Blue}: best performance with statistical significance within each method category. Statistical significance level: $p < 0.05$.} 
\label{tab:extvalid_results}

\end{table*}

\section{Experiments} \label{sec:exp}

%\hwnote{we need initial introduction for each section.}
In this section, we describe the baseline and prior methods that we compare to. We also provide experimental protocol and evaluation metric details.

\noindent\textbf{Baselines} We compare our proposed model to several commonly used strategies for imputing missing values in time series data, as well as to the state-of-the-art imputation method proposed by \cite{jang2020deep}. We group methods into several categories. Simple filling methods include zero fill, forward fill, backward fill, the average of them (Avg.F+B), mean fill, micro mean fill and median fill. Here, mean filling uses the mean of the hourly step count computed over a specified set of hourly blocks while micro mean filling uses the total step count divided by the total wear time where the totals are computed over a specified set of hourly blocks. 

The mean, micro mean and median based methods are applied in four variations corresponding to computing the imputation statistic over different sets of hourly blocks. All are applied on a per-participant basis. For example, in the ``Participant" variant we compute a per-participant imputation statistic over all available data for a single participant and then apply it to all missing hourly blocks for that participant. In the ``DW+HD" variant, we compute an imputation statistic per hour of the day and day of the week for each participant and apply it to all missing data from that hour of day and day of week combination for that participant.

The $k$NN model includes two variants: uniform, which assigns uniform weights to neighbors, and softmax, where weights depend on an RBF kernel based on the distances between the target hourly block and its neighbors. 
Finally, model-based baseline methods include  linear regression imputation, iterative imputation (which iteratively estimates variables with missing values from other observed variables \citep{azur2011multiple}), the stat-of-the-art convolutional denoising autoencoder (CNN-DAE) model of \cite{jang2020deep}, the RNN models BRITS \citep{cao2018brits} and MRNN \citep{yoon2018estimating}, the USGAN model of \cite{miao2021generative}, and the attention model SAITs \citep{du2023saits}. 

\noindent\textbf{Handling Missing Input Features} Multiple models that we consider including basic regression imputation and the proposed model will have missing values in their input feature representations. 
We address missing data in the LAPR feature representation using DW+HD median imputation. This choice is made since DW+HD median filling is the most accurate of the basic imputation methods on these data and often outperforms $k$NN imputation. For mean and median imputation methods, if there are no observed hourly blocks associated with a specific hour of the day or day of the week, we apply participant-level median imputation to all the hourly blocks associated with that particular hour of the day or day of the week. For more information on how we handle feature missingness in other baseline models, please refer to Appendix \ref{sec:model_setups}.

\begin{figure}[t]
\centering
\includegraphics[height=4cm, width=.6\textwidth, keepaspectratio]{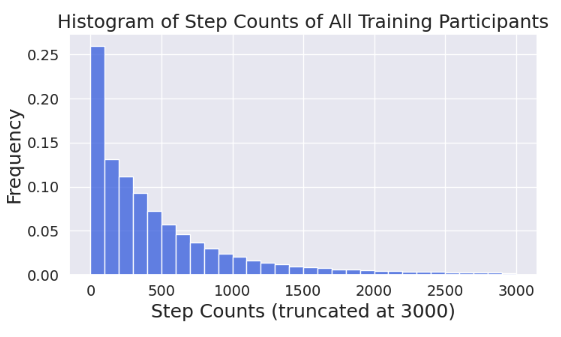}
\caption{Histogram of observed hourly step counts between 6:00am and 10:00pm for the 100 training participants.}
\label{fig:hist_step_count}
\end{figure}

\noindent\textbf{Data Partitioning} The proposed model and multiple baseline approaches include hyper-parameters that need to be set. To accomplish this, we apply a 10-fold stratified random sampling validation approach to the training data set described in Section \ref{sec:data}. We use a stratified approach because the target step count variable is significantly skewed toward low step count values as seen in Figure \ref{fig:hist_step_count}. When holding out instances, it is thus important to match these statistics since an over or under abundance of large step count values can have a large effect on validation set performance estimates. We use per-participant uniform density bins in the stratified sampling. In terms of the data partitioning scheme, we allocate 80\% of instances in each split for training, 15\% for validation and 5\% for an in-domain test set. However, in this work we focus on the fully held out test set described in Section \ref{sec:data} to provide results covering multiple levels of missing data.

\noindent\textbf{Hyper-Parameter Optimization} The stratified train/validation splits are used to select hyper-parameters for all $k$NN-based and model-based approaches including the proposed model. Details including model configurations, selected hyper-parameters and full training procedures can be found in Appendix \ref{sec:model_setups}.

\noindent\textbf{Model Evaluation} We evaluate trained models on the completely held out test set as described in Section \ref{sec:data}. Results are reported per missing data bin as well as overall. We report results in terms of Macro Mean Absolute Error (MAE). This is the mean over participants in the test set of the mean absolute error per test participant, which is defined in Equation \ref{equ:macro_mae}. 

\begin{equation}
\label{equ:macro_mae}
    \textit{Macro MAE} = \frac{1}{N} \sum_{n=1}^{N} \frac{1}{|\mathcal{M}^{(n)}|} \sum_{m_n=1}^{|\mathcal{M}^{(n)}|} AE_{m_n}
\end{equation}

\noindent where $m_n \in \mathcal{M}^{(n)}$ is the index of a single hourly block to be imputed from the set of missing hourly blocks $\mathcal{M}^{(n)}$ of participant $n$. $N$ is the number of participants in the dataset and $|\mathcal{M}^{(n)}|$ is the number of imputed hourly blocks from participant $n$.

As a measure of variation, we report $\pm 1.96$ times the standard error of the mean, yielding a 95\% confidence interval on mean predictive performance. For models where hyper-parameters are selected using the 10 validation splits, we determine the optimal hyper-parameter values using the validation set and average the test predictions of the 10 corresponding models to form a final test prediction. For personalized baseline models (e.g., participant-level mean imputation), we use imputation statistics computed from the test data set. This is necessary because these approaches are applied per-person and the test set consists of completely held-out individuals with no overlapping data in the training set. This biases these results in favor of the baselines.

%%%% RESULTS %%%%

%%%%%%%%% Model Comparison (Hourly Level) %%%%%%%%%%%%%%%%%
\begin{figure*}[htp!]
\centering
\includegraphics[width=\linewidth]{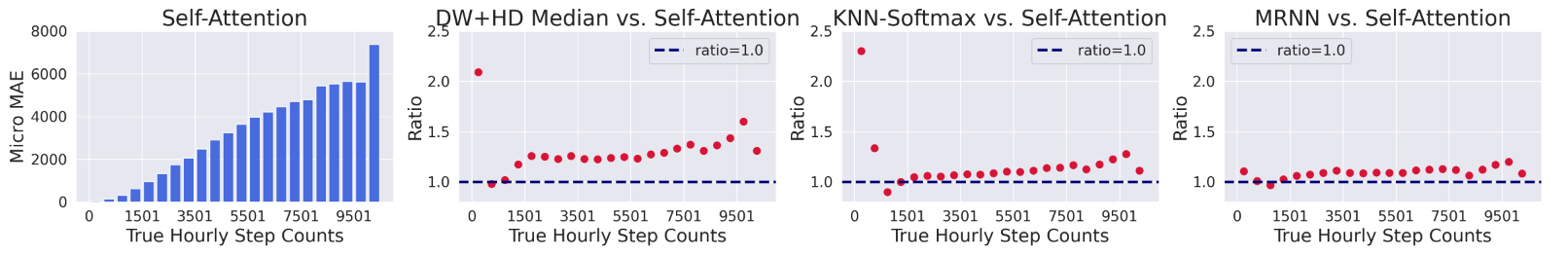}
\caption{Imputation results and model comparison on hourly blocks with various ground truth step counts. The first plot shows the proposed model's performance (evaluated by Micro MAE) by true step count bins. The first bin is for zero steps, while the rest have the bin width of 500 steps (i.e., [1, 500), [501, 1000), etc). The second to fourth plot show error ratios relative to the proposed model for several other models. Error ratios above 1 indicate that other models perform worse than the proposed model on the particular bin.}

\label{fig:model_compare_hourly}
\end{figure*}

\begin{figure*}[t]
\centering
\includegraphics[width=\linewidth]{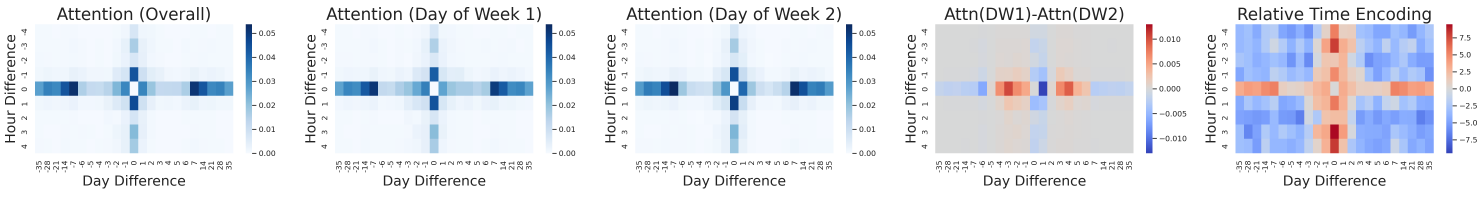}
\caption{Attention and relative time encoding visualization. We include attention weights regarding two days of the week as examples, and also show the attention difference between them (the fourth image). The attention scores are averaged over all completely held-out test samples, and relative time encoding is averaged over models from 10 training splits.}
\label{fig:attn_enc}
\end{figure*}

\section{Results} \label{sec:results}
In this section, we present step count imputation results on the 500-participant test set. Further, we visualize the attention maps and relative time encodings learned by the proposed model to analyze what the model learns from data. Finally, we provide the results of an ablation study varying components used in the self-attention model.

\noindent\textbf{Overall Imputation Results} \tableref{tab:extvalid_results} shows the overall imputation results (last column) for each method. Methods highlighted in blue have statistically significantly lower error than other methods in their group ($p<0.05$). Methods highlighted in red have statistically significantly lower error than other methods across all groups ($p<0.05$). As we can see, our sparse self-attention model achieves the best overall performance and does so with statistical significance relative to all other methods considered.

\noindent\textbf{Imputation Results by Missingness Rate} The remaining columns in \tableref{tab:extvalid_results} show the imputation results for each missingness rate interval. As we can see, our sparse self-attention model achieves the best performance on all but the highest missing data rate bin. On participants with extremely high missing rates (i.e. $\geq 80\%$), DW+HD Median Fill performs best and is better than our self-attention model with statistical significance ($p<0.05$). This is likely due to the fact that at over 80\% missing data, the context windows for the proposed model will contain relatively few observations while the LAPR feature vectors will be heavily influenced by the baseline imputation method used. It may be possible to further improve performance for high missing rate bins by using adaptive context window sizes and alternative LAPR construction methods or by adaptively smoothing the model's prediction towards that of simpler models as the volume of observed data in the context window decreases.\footnote{We note all 95\% confidence intervals reported in the table represent $\pm 1.96$ times the standard error of the mean MAE for each model. These intervals are wide due to variability across participants in our dataset. However, the paired t-test depends instead on the distribution of per-participant \emph{differences} in performance between two models.}

\noindent \textbf{Imputation Results by Step Count} We further analyze the imputation results by breaking the overall performance down based on ground truth step count bins for different models. The performance is evaluated in terms of micro MAE per ground truth step-count bin. The first plot in \figureref{fig:model_compare_hourly} shows the test error rate of the proposed model per ground truth step count bin. We can see that the model has higher error on bins corresponding to higher ground truth step counts. This is perhaps not surprising as high ground truth hourly step counts occur much more rarely than low step counts as seen in Figure \ref{fig:hist_step_count}. The remaining plots in \figureref{fig:model_compare_hourly} present the ratio of the error obtained by the DW+HD Median, $k$NN-Softmax and MRNN approaches (the best other models in their groups) to that obtained by the proposed model. Ratios above 1 indicate that the alternative models have higher error than the proposed model. We can see that the proposed model not only outperforms the alternative models overall, it does so with respect to almost all individual ground truth step count bins. 

\noindent\textbf{Attention and Relative Time Encoding Visualization} Figure 
\ref{fig:attn_enc} shows the attention weights averaged over all instances, the attention weights averaged over specific example days, and  the relative time encodings. From these visualizations, we can see that the model produces overall average attention weights that match expectations based on the  autocorrelation function shown in Figure \ref{fig:acf}. The time points with consistently high attention relative to the target hour are $\Delta 
t=\pm 1 \text{ hr}, \pm 1 \text{ day}, \pm k \text{ weeks}$. Further, we can see that the average attention weights are not the same for all days of the week. The model produces different average attention weights for different days. Lastly, we can see clear difference between the relative time encoding structure and the overall average attention weights 
thus clearly indicating that both the input features and the relative time encoding influence the attention weights. 

%Attention weights for all days are included in Appendix 
%\ref{sec:attn_alldayweeks}.

\begin{figure}[t]
\centering
\includegraphics[height=2.5cm, width=\columnwidth, keepaspectratio]{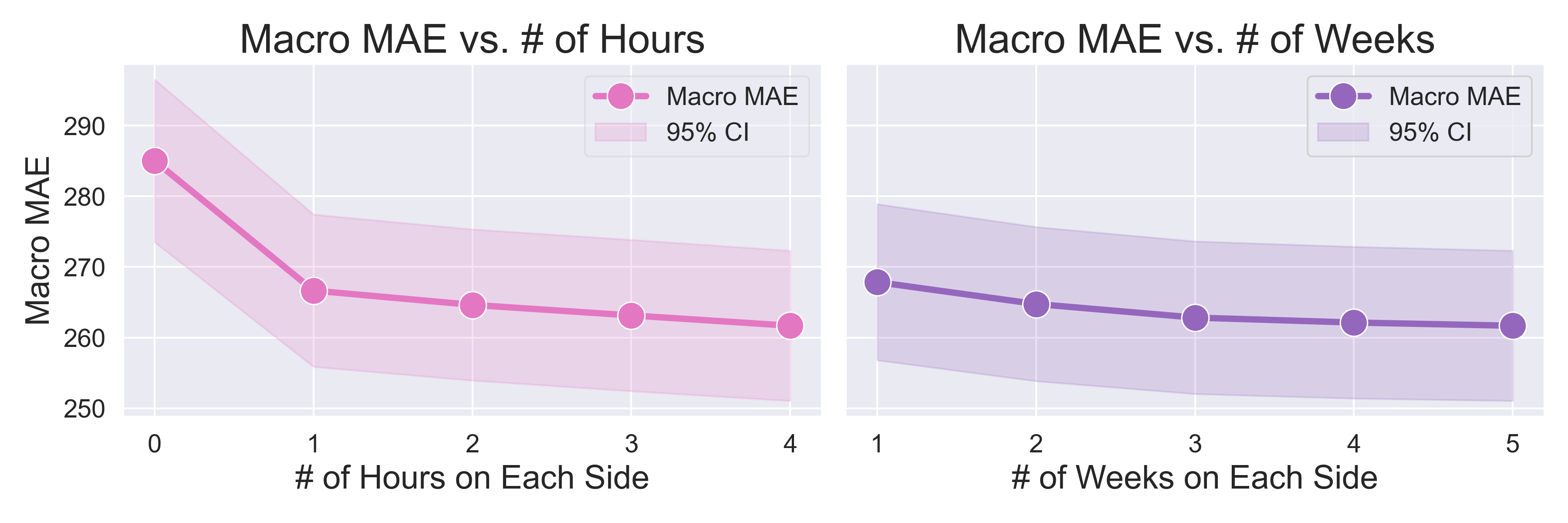}
\caption{Ablation results on the local context window size including varying the number of hours (left) and number of week (right).}
\label{fig:ablation_hours_weeks}
\end{figure}

%\subsection{Ablation Study}
\noindent\textbf{Ablation Study} We conduct an ablation study to test the impact of local context window sizes and different architecture components used in our sparse self-attention model. Macro MAE of the held-out test samples is used to measure performance. We first consider the effect of changing both the number of weeks represented in the context window and the number of hours. The results are shown in \figureref{fig:ablation_hours_weeks}. We see that as the number of weeks and the number of hours is increased, the prediction error decreases. These results support the importance of using wider context windows spanning multiple weeks. The model used in the main results corresponds to hours=4 and weeks=5. 

We next consider the impact of the relative time encoding and local activity profile representation (LAPR). Removing the relative time encoding increases the overall test error from 261.68 ± 10.62 to 262.91 ± 10.75. While the error increases, the increase is not statistically significant. When removing the LAPR from the model's input features, the error increases to 278.75 ± 11.93. This increase in error is significant, indicating that the LAPR provides a valuable performance boost to the model relative to using the base features associated with each hourly block.

%%%% CONCUSION %%%%
\section{Conclusions} \label{sec:conc}
% Wearable physiological sensors like those found in smartwatches are pivotal for improving the health outcomes of individuals with chronic conditions such as heart disease and obesity. However, the inherent complexity and frequency of missing data patterns within wearable sensor data pose significant challenges for researchers seeking actionable insights into health behaviors, especially in the case of physical activity data with its substantial variabilities. We introduce the \textit{StepImpute} challenge, featuring a representative cohort from the \textit{All of Us} archive, to encourage the machine learning community to address the challenge of extracting meaningful activity patterns from complex missing data and facilitating health interventions. Our contributions include an extraction pipeline for creating a representative dataset and a novel imputation method using sparse self-attention for reconstructing missing step count data, thus setting the stage for future research in this domain.

In this work, we consider the problem of imputing missing step count data collected by wearable devices. To enable this research, we curated a novel dataset consisting of 100 training participants and 500 test participants with more than 5.5 million total hourly step count observations extracted from the \textit{All of Us} dataset. We proposed a customized model for this task based on a novel multi-timescale sparse self attention structure to mitigate the quadratic  complexity of the standard dense self-attention mechanism. 

Our experiments show that the proposed model outperform the considered baseline approaches and prior state-of-the-art CNN-based models on fully held out test data. Further, we present ablation studies showing the importance of both the activity profile input representation that we propose and the multi-timescale attention computation. We note that although our model and feature representations were specifically designed for step-count data in this paper, the same structures could also be helpful for modeling other behavioral and physiological processes (e.g. heart rates) with similar quasi-periodic and multi-timescale structures across day, weeks and months.

In terms of limitations, we first note that computational considerations limited the total data volume that could be used for model training in this work. While we opted to use a training data set containing fewer participants with higher observed data rates, designs using randomly selected training participants with similar total training data volume would also be feasible. 

While there may be concern that the training set is not representative of the data set over all, the test set is indeed a fully held out and representative stratified random sample and the proposed model achieves superior overall performance on this test set. Next, we note that the missing data mechanism used when evaluating models is effectively a missing completely at random (MCAR) mechanism. However, the per-step count results presented in Figure \ref{fig:model_compare_hourly} provide information about the distribution of predictive performance conditioned on true step counts. 

In terms of future work, we plan to extend the proposed model to a multi-layer architecture to mitigate the fact that the input feature representation relies on simple imputation currently. Applying the model in multiple layers may further improve performance by providing more accurate local activity profile representations. In addition, we plan to extend the model to produce probabilistic predictions to support multiple imputation workflows and to extend the model architecture to several related tasks including step count and sedentary interval forecasting.
Finally, we plan to evaluate the impact of the imputations produced by the model when applied as part of a data analysis procedure that aims to quantify the association between physical activity as measured by step count data and a related health condition or intervention outcome. 
%Of particular interest is a sensitivity analysis of association levels as a function of parameters of the imputation process. 

%%%% ACNOWLEDGEMENT %%%
\section*{Acknowledgments}
This work was partially supported by National Institutes of Health National Cancer Institute, Office of Behavior and Social Sciences, and National Institute of Biomedical Imaging and Bioengineering through grants U01CA229445 and 1P41EB028242 as well as by a Google Cloud Research Credits Program credit award. We gratefully acknowledge \textit{All of Us} participants for their contributions, without whom this research would not have been possible. We also thank the National Institutes of Health’s \textit{All of Us} Research Program for making available the participant data examined in this study.

%%%% REFERENCE %%%%
\bibliography{citations}

%%%% APPENDIX %%%%
\appendix
\clearpage

\section{Data Curation and Preprocessing Pipeline} \label{sec:curation_preprocess}
\begin{figure}[ht!]
\centering
\includegraphics[width=\columnwidth]{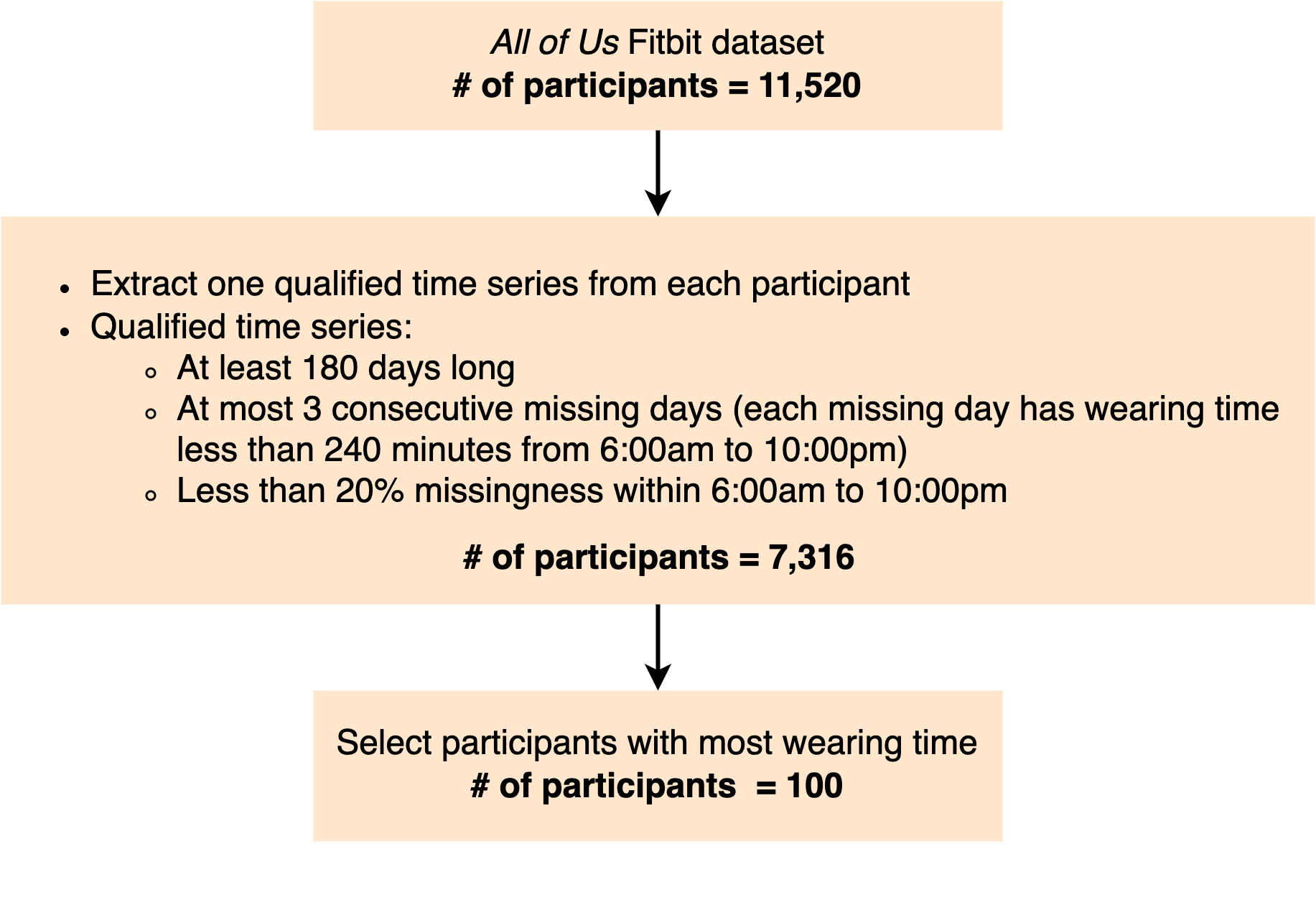}
\caption{Flow chart for cohort curation}
\label{fig:cohort_describe}
\end{figure}

\begin{figure}[!htb]
\centering
\includegraphics[width=\columnwidth]{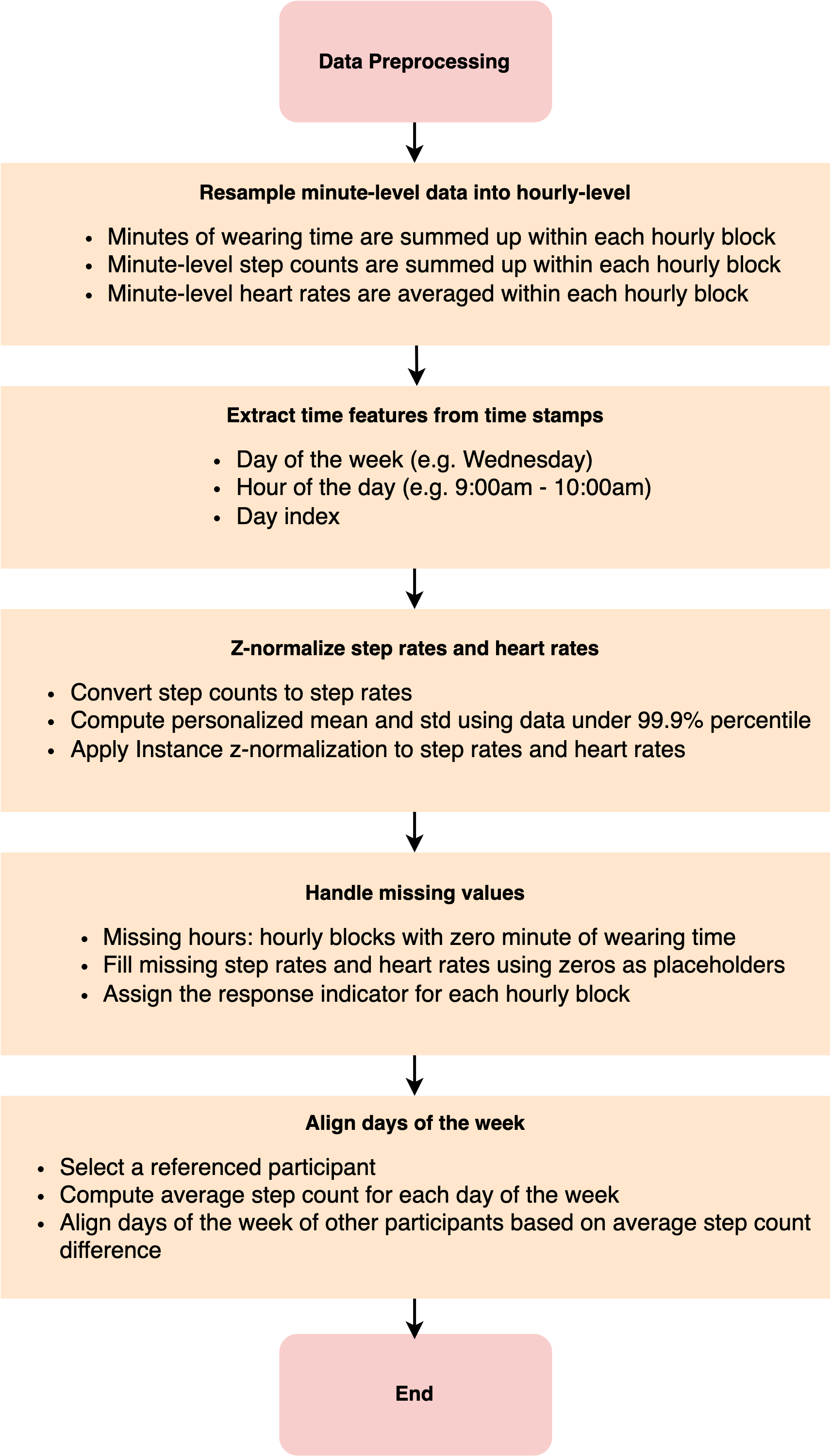}
\caption{Data preprocessing pipeline}
\label{fig:data_preprocessing}
\end{figure}

Figure \ref{fig:cohort_describe} and Figure \ref{fig:data_preprocessing} demonstrates how we curate the training cohort and preprocess the data. 

%\clearpage

\section{Comparison between Training Cohort and All \textit{All of Us} Participants} \label{sec:comp_cohort_allofus}
Figure \ref{fig:stats_basic} and \ref{fig:stats_hr_sr_miss} compare the statistics of the training cohort of 100 participants with the entire \textit{All of Us} Fitbit dataset of 11,520 participants. 

\begin{figure*}[htbp!]
\floatconts
  {fig:stats_basic}
  {\caption{\textbf{Blue}: Statistics of the entire \textit{All of Us} Fitbit dataset with 11,520 participants. \textbf{Orange}: Statistics of the curated training cohort with 100 participants. \textbf{First column}: Distribution of the total number of hourly blocks from each participant. \textbf{Second column}: The total number of valid hourly blocks (i.e. hourly blocks with non-zero wearing minutes) from each participant. \textbf{Third column}: Missing rate (i.e. the number of invalid hourly blocks divided by the total number of hourly blocks), including all hours of the day in each participant, not only from 6:00am to 10:00pm.}}
  {%
    \subfigure[\textit{All of Us} dataset]{\label{fig:entire_basic}%
      \includegraphics[width=\linewidth]{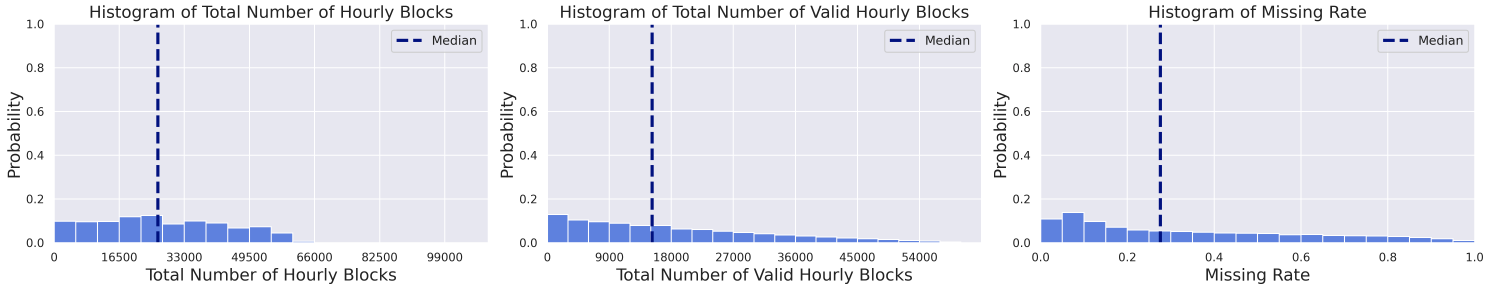}}%
    \qquad
    \subfigure[Training cohort]{\label{fig:top_basic}%
      \includegraphics[width=\linewidth]{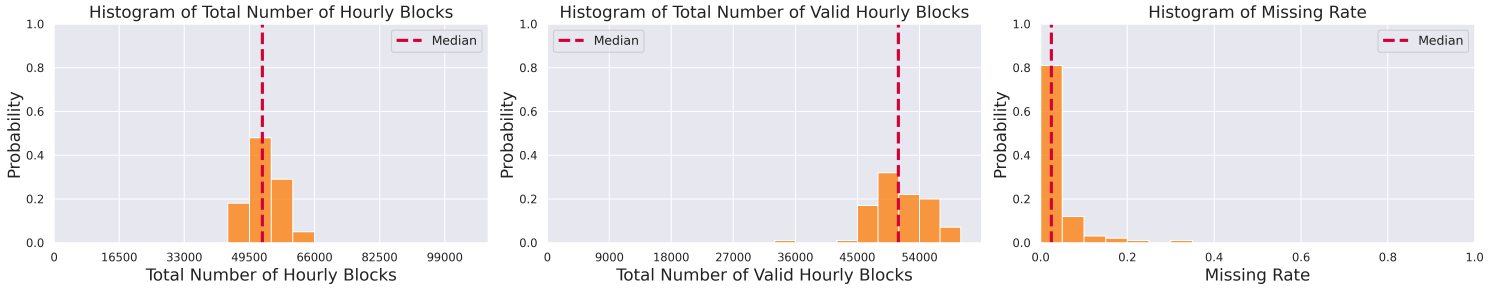}}
  }
\end{figure*}

\begin{figure*}[htbp!]
\floatconts
  {fig:stats_hr_sr_miss}
  {\caption{\textbf{Blue}: Statistics of the entire \textit{All of Us} Fitbit dataset with 11,520 participants. \textbf{Orange}: Statistics of the training cohort with 100 participants. \textbf{First column}: Distribution of average heart rates over all the participants. \textbf{Second column}: Distribution of average hourly step rates over all the participants. \textbf{Third column}: Distribution of hourly step rates of each hour of the day. \textbf{Fourth column}: Missing rate of each hour of the day. The average heart rates and average step rates are computed over all the observed hourly blocks for each participant.}}
  {%
    \subfigure[\textit{All of Us} dataset]{\label{fig:entire_hr_sr_miss}%
      \includegraphics[width=\linewidth]{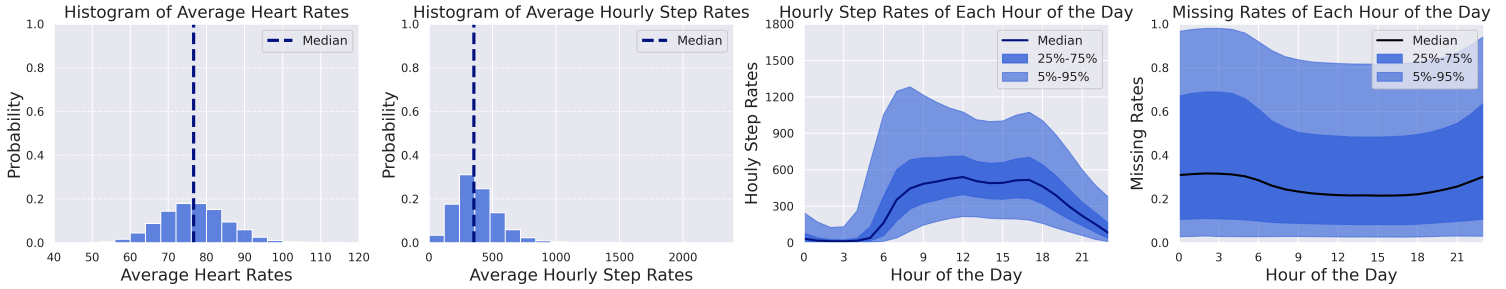}}%
    \qquad
    \subfigure[Training cohort]{\label{fig:top_hr_sr_miss}%
      \includegraphics[width=\linewidth]{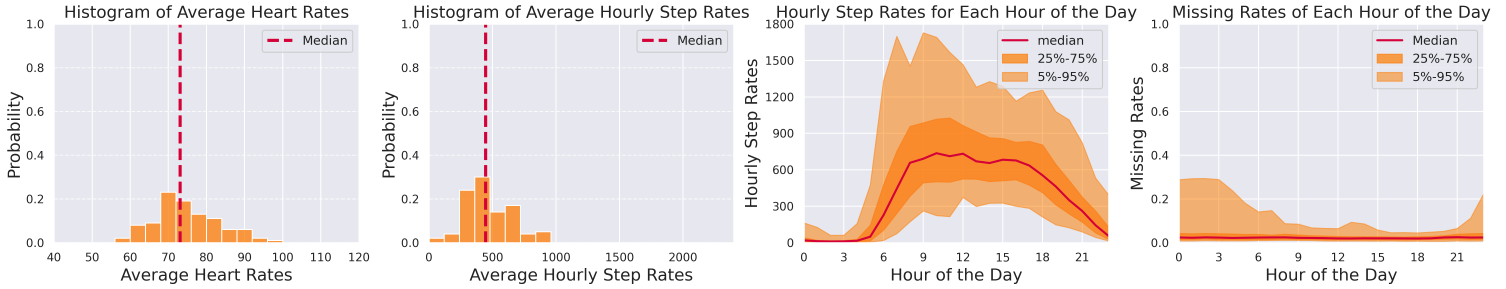}}
  }
\end{figure*}

\section{Model Configurations, Hyper-Parameters and Training Procedures} \label{sec:model_setups}
In this section, we introduce the details about all the models used in our experiments, including configurations, hyper-parameters and training procedures.  

\subsection{Multi-Timescale Sparse Self-Attention Model}
We fix the length of local activity profile representations (LAPR) as $2W+1 = 2\times72+1 = 145$. The configuration of the LAPR encoder network is: $Conv$: out\_channels=1, kernel\_size=49, stride=1, padding=24, with no bias; $Average Pool$: kernel\_size=7 and stride=6. 

The model is trained using Adam optimizer with the batch size of 20,000 for 30 epochs. The learning rate is searched within \{0.1, 0.01, 0.001\}. We conduct early stopping based on validation Micro MAE for each split. Validation Micro MAE averaged over 10 splits is used to choose the best hyper-parameters. We train the model using two NVIDIA Tesla T4 GPUs with 32 CPUs and 208 GB RAM within \textit{All of Us} workspace. The model is implemented using PyTorch 1.13.1.

\subsection{Filling Methods} 
All the filling methods impute missingness on the level of \emph{unnormalized} step rates (i.e. before instance z-normalization). Micro mean, mean and median based methods compute statistics of all levels (e.g. participant level) using the data from 6:00am to 10:00pm, while Forward and Backward Fill based methods are allowed to use the data out of this period.

\subsection{Regression Imputation}  
We set the regression function to be linear. Input features of the linear regression model include (1) normalized step rates and heart rates from all the blocks in the context window, except for the center one (2) day of the week and hour of the day one-hot vectors of the center hourly block. LAPR is not applied as it was found to decrease performance. Missing step rates and heart rates is filled by zeros, which exhibits superior performance compared to DW+HD median filling. The model has the same context window size, training protocol and loss function as our proposed model. We set the batch size as 50,000 and search for the learning rate within $\{0.1, 0.01, 0.001, 0.0001\}$. Adam optimizer is used train the model for 20 epochs with the learning rate of 0.001.

\subsection{$k$-Nearest Neighbors ($k$NN) Imputation}
We search for nearest neighbors within all the observed data of the same participant where the missing block comes from. The neighbors are \emph{not} limited to 6:00am to 10:00pm period. Input features are LAPR with the same length (i.e.,145) used in the proposed model. Two variations are tested: (1) uniform weighting ($k$NN-Uniform) and (2) RBF-kernel-based method ($k$NN-Softmax), where the similarity between the missing hourly block and its neighbors depends on square distances in the feature space. We search for the number of nearest neighbors in \{1, 7, 14, 21, 28, 35\} for both and the RBF parameter within \{0.1, 0.01, 0.001, 0.0001, 0.00001\} for $k$NN-Softmax.

\subsection{Multiple Iterative Imputation Method (Iterative Imputation)}
\begin{figure}[ht]
\centering
\includegraphics[width=0.8\columnwidth]{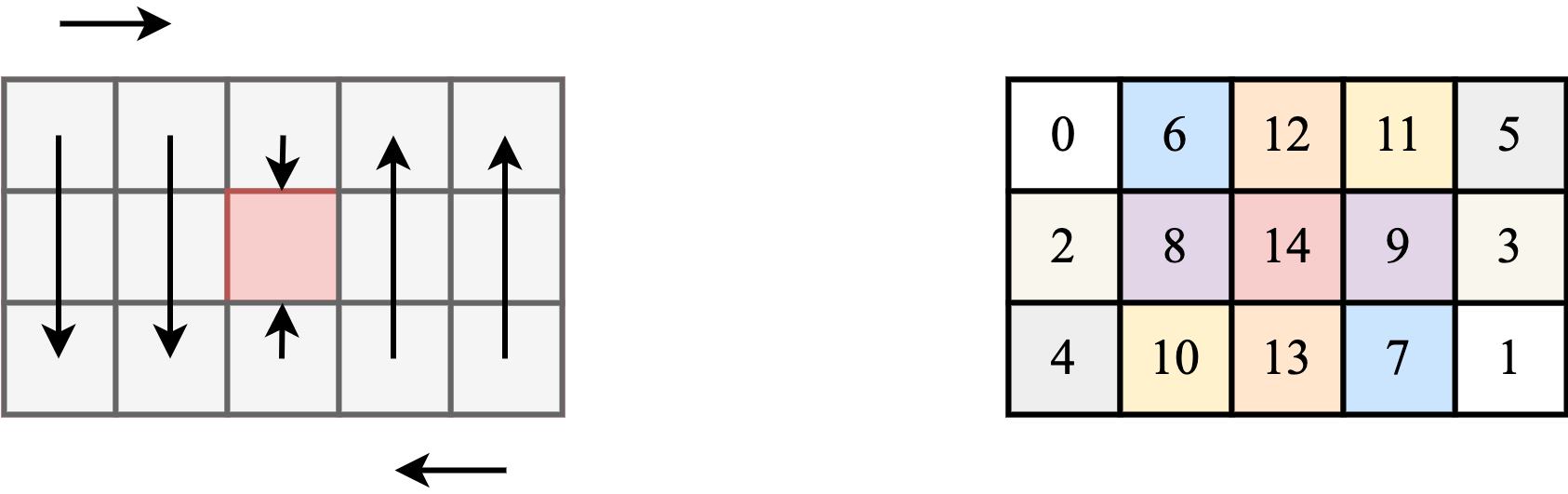}
\caption{The order of prediction positions within the context window using iterative imputations. The order is alternating between the start (i.e. upper left) and the end (i.e. lower right). The direction from the start is from top to bottom and from left to right, while the direction from the end is from bottom to top and from right to left. The plot on the right gives an example. The number means the order of computation and the same color means the same relative position from the start and from the end.} 
\label{fig:mice_order}
\end{figure}

Our model is similar to Multiple Imputation with Chained Equations (MICE) method, which uses chained equations and linear regression models to impute every variable conditioned on the others. However, during the training phase, the algorithm performs a deterministic imputation instead of probabilistic sampling. The input features are the same as used for regression imputation. Since day of the week and hour of the day are always observed, they only serves as the input features while imputing other variables, and themselves are never imputed.  Figure \ref{fig:mice_order} provides an example of our specified imputation order regarding positions in the contex window. Each linear regression model in the chained equation is trained using mini-batch SGD with the batch size of 10,000 for 2 epochs. The number of imputation iterations is set as 2. During inference, we perform multiple imputations for each position by sampling from a Gaussian distribution. Please refer to the codes for the details. 

\subsection{Convolutional Denoise Autoencoder (CNN-DAE)}
We use the symmetric encoder-decoder architecture to implement CNN-DAE. The encoder consists of three 1D convolutional layers, each followed by BatchNorm and ReLU activation. Correspondingly, the decoder includes three 1D transposed convolutional layers, with the first two layers being followed by Batch Normalization and ReLU activation. The configurations of convolutional and transposed convolutional layers are in Table \ref{tab:conv_config}. For the input features, we include both z-normalized step rates and heart rates within the same context window used in the proposed methods. Since the CNN model structure is not suitable for use with the multi-scale context window, we apply it at the hourly level to the contiguous time span. Furthermore, LAPR is not employed in CNN-DAE due to its inability to yield better performance. We fill the missingness with zeros. Adam optimizer with batch size as 50,000 is used to train the model of each split for 20 epochs. The learning rate is searched within \{0.1, 0.01, 0.001\}.

\begin{table}[h!]
\caption{Configurations of Convolutional and Transpose Convolutional Layers in CNN-DAE}
\small
\centering
\resizebox{\columnwidth}{!}{%
\begin{tabular}{@{}llllll@{}}
\toprule
\textbf{Layer} & \textbf{Input Channel} & \textbf{Output Channel} & \textbf{Kernel Size} & \textbf{Stride} & \textbf{Padding} \\ \midrule
Conv1      & 2  & 4  & 31 & 2 & 11 \\
Conv2      & 4  & 8  & 20 & 2 & 9  \\
Conv3      & 8  & 16 & 10 & 2 & 4  \\ \midrule
TransConv1 & 16 & 8  & 10 & 2 & 4  \\
TransConv2 & 8  & 4  & 20 & 2 & 9  \\
TransConv3 & 4  & 2  & 31 & 2 & 11 \\ \bottomrule
\end{tabular}%
}
\label{tab:conv_config}
\end{table}

\subsection{BRITS} \label{sec:brits}
We adhere to the settings in the original paper, using LSTM as the RNN architecture. Input features are the same as the proposed model.However, we found LAPR cannot help to improve the performance, thus we did not use it here. The context window is chronologically flattened, enabling the RNN model to process information sequentially. We impute both heart rates and step rates at each time step. Notebly, we found that the auxiliary heart rate imputation task indeed helps the step rate imputation task for BRITS, so we keep both of them during training. The best hyper-parameters are selected based on the optimal validation Micro MAE of step counts of the center hourly blocks. Training the BRITS model spans 30 epochs with the batch size of 10,000 and the learning rate of 0.01. The LSTM hidden dimension is searched within \{4, 8, 16, 32\}.

\subsection{MRNN}
MRNN consists of the interpolation block and the imputation block. In the interpolation block, we apply two bidirectional-GRU models to interpolate the missing values, one for step rates and the other for heart rates. 
Day of the week (DW) and hour of the day (HD), which are always observed, are not input into the interpolation block since it operates within each data stream with missing values. On the contrary, they are input to the imputation block. The context window is consistent with that used in the proposed model. As suggested by the original paper, missing values outside of the center hourly block are filled with zeros. We found DW+HD median filling does not demonstrate the performance as good as zero-filling. Like BRITS, the context window is flattened in chronological order for RNN to process. We also found LAPR can improve the MRNN performance as with our proposed model, thus these features are used when reporting the results. To keep consistent with other models, we employ Mean Absolute Error (MAE) instead of Mean Squared Error (MSE) for model training, different from the original paper. We train MRNN for 40 epochs, utilizing the batch size of 20,000 and the learning rate of 0.01. The GRU hidden dimension in the interpolation block is searched within \{4, 8, 16, 32\}.

\subsection{USGAN}
We employ the BRITS model as the generator and the bidirectional GRU model as the discriminator. The generator configurations align with those outlined in \sectionref{sec:brits}. As our data does not have explicit labels for each time series, we omit the classifier component mentioned in the original paper. In contrast to the original implementation, which updates the discriminator five times after each generator update, updating the discriminator only once results in more stable training and improved performance in our case. We train the USGAN model for 30 epochs with the batch size of 10,000 and the learning rate of 0.01. The RNN hidden dimensions for both the generator and discriminator are explored within \{4, 8, 16, 32, 64\}. Additionally, we search for the weight of the discriminator loss during training, which balances it with the BRITS loss, within \{0.1, 0.3, 0.5, 0.7, 0.9, 1.0\}.

\subsection{SAITS}
We use the learning rate of 0.01 and the batch size of 10,000 when training the model. We fix the number of transformer layers as 2 and search for the hidden representation dimension $d_{model}$ and the output dimension of each layer $d_v$ within \{4, 8, 16, 32\}. We leveraged the same multi-scale context window\footnote{We note that vanilla SAITS model uses the dense self-attention, which is not feasible in our case due to the long time series data.} as in our proposed model as well as the same feature set, including the LAPR.

\end{document}